%% file: main.tex
\pdfoutput=1

\documentclass[11pt]{article}

\usepackage[final
]{acl}

\usepackage{times}
\usepackage{latexsym}
\usepackage{enumitem}
\usepackage{booktabs}
\usepackage{float}
\usepackage{subcaption}

\usepackage[T1]{fontenc}

\usepackage[utf8]{inputenc}

\usepackage{microtype}

\usepackage{inconsolata}

\usepackage{amsmath}
\usepackage{amsfonts}
\usepackage{bbm}
\usepackage{commath}
\usepackage{comment}
\usepackage{amsthm}   
\usepackage{hyperref} 
\usepackage{cleveref} 
\usepackage{lipsum}   
\usepackage{subcaption}
\usepackage{wrapfig}
\usepackage{multirow}

\makeatletter
\newcommand\nocaption{%
    \renewcommand\p@subfigure{}
    \renewcommand\thesubfigure{\thefigure\alph{subfigure}}}
\makeatother

\makeatletter
\newtheorem*{rep@theorem}{\rep@title}
\newcommand{\newreptheorem}[2]{%
\newenvironment{rep#1}[1]{%
 \def\rep@title{#2 \ref{##1}}%
 \begin{rep@theorem}}%
 {\end{rep@theorem}}}
\makeatother

\newtheorem{lemma}{Lemma}

\newreptheorem{lemma}{Lemma}

\DeclareMathOperator*{\argmin}{arg\,min}

\usepackage{listings, lstautogobble}
\usepackage{graphicx}

\definecolor{midnightgreen}{rgb}{0.0, 0.29, 0.33}

\title{On the Feasibility of In-Context Probing for Data Attribution}

\author{
    Cathy Jiao\textsuperscript{1} \quad 
    Weizhen Gao\textsuperscript{2} \quad 
    Aditi Raghunathan\textsuperscript{2} \quad 
    Chenyan Xiong\textsuperscript{1} \\  
    \textsuperscript{1} Language Technologies Institute, Carnegie Mellon University \\
    \textsuperscript{2} Computer Science Department, Carnegie Mellon University \\
\texttt{\small \{cljiao, aditirag, cx\}@cs.cmu.edu, \small{wgao2@andrew.cmu.edu}}}

\begin{document}
\maketitle

\input{sections/0_abstract}
\input{sections/1_introduction}

\input{sections/2_related_work}
\input{sections/3_background}

\input{sections/4_setup}
\input{sections/5_experiments}

\input{sections/6_data_selection_experiments}
\input{sections/7_synthetic_experiments}
\input{sections/8_conclusion}
\input{sections/ethics_limitations}
\input{sections/acknowledgements}

\bibliography{anthology,custom}
\newpage

\appendix
\input{sections/appendix}

\end{document}

%% file: sections/0_abstract.tex
\begin{abstract}
Data attribution methods are used to measure the contribution of training data towards model outputs, and have several important applications in areas such as dataset curation and model interpretability. However, many standard data attribution methods, such as influence functions, utilize model gradients and are computationally expensive. In our paper, we show in-context probing (ICP) -- prompting a LLM -- can serve as a fast proxy for gradient-based data attribution for data selection under conditions contingent on data similarity. We study this connection empirically on standard NLP tasks, and show that ICP and gradient-based data attribution are well-correlated in identifying influential training data for tasks that share similar \textit{task type} and \textit{content} as the training data. Additionally, fine-tuning models on influential data selected by both methods achieves comparable downstream performance, further emphasizing their similarities. We also examine the connection between ICP and gradient-based data attribution using synthetic data on linear regression tasks. Our synthetic data experiments show similar results with those from NLP tasks, suggesting that this connection can be isolated in simpler settings, which offers a pathway to bridging their differences. \footnote{Code/data at \url{https://github.com/cxcscmu/InContextDataAttribution}}
\end{abstract}

%% file: sections/1_introduction.tex
\section{Introduction}\label{sec:introduction}

Data attribution methods aim to identify specific training data that contribute to the outputs of a model \citep{worledge2023unifying}. Methods for data attribution have numerous useful applications; for instance, dataset curation \citep{ilyas2022datamodels, xia2024less, yu2024mates}, model interpretability \citep{han2021influencetuningdemotingspurious, akyürek2022tracingfactualknowledgelanguage, li2024attributionbenchhardautomaticattribution}, and data valuation \citep{ghorbani2019datashapley, yang2024gmvaluatorsimilaritybaseddatavaluation, zhang2025fairsharedatapricinglarge}. While data attribution methods are useful, they can be computationally demanding. For instance, influence functions \citep{koh2017understanding} are a classic tool for gradient-based data attribution (i.e., methods that utilize model gradients in their computation), but are challenging to scale for large deep learning models with billions of parameters \citep{grosse2023studying, choe2024dataworthgptllmscale}.

Recently, data selection using in-context probing (ICP) -- prompting a LLM -- to determine the quality of a training data sample has become an important avenue for curating high-quality training datasets \citep{rubin2022learning, nguyen2023incontext, wettig2024qurating}. Yet, it is unclear why ICP is effective at training data selection since there are multiple factors to consider for determining the quality of training data, such as mixtures, utility, and the quantity of data \citep{lee2022deduplicating, xie2023doremi, goyal2024scaling}. 
 
In this paper, we offer an explanation for this phenomenon by drawing a connection between ICP and gradient-based data attribution. To study the robustness of this connection, we empirically analyze the agreement between both methods for identifying influential training data for in-domain target tasks (i.e., tasks that share similar \textit{task type} and \textit{content} as the training data), and out-of-domain target tasks. On standard NLP tasks (including instruction-following and QA), our experiments reveal that ICP can approximate gradient-based data attribution for identifying influential training data in the in-domain setting. Further fine-tuning on the influential training data selected by either method — in particular, using data from the Alpaca Dataset \citep{taori2023stanford} — results in similar model performance in instruction-following on Alpaca Eval \citep{li2023alpacaeval, dubois2024alpacafarm}. This is advantageous since, unlike gradient-based attribution methods, ICP enables cost-effective data selection; it requires no access to model parameters, and can even be performed via API calls, making it ideal for black-box models.

In addition to standard NLP tasks, we study the connection between ICP and gradient-based data attribution in a controlled setting using synthetic data, specifically linear regression tasks. In this setting, the \textit{task type} (i.e., the specifically linear relation) and \textit{content} (i.e., input distance) of the training data and target task are clearly defined, making them easy to adjust. Similar to standard NLP tasks, our findings on synthetic data show that ICP can approximate gradient-based data attribution in the in-domain setting. Furthermore, our synthetic data results show that this connection can be isolated, which paves way for future research bridging the gap between the two methods. Our contributions are summarized as follows:

\begin{enumerate}
    \item We draw a connection between ICP and gradient-based data attribution and show they agree in identifying influential training data for in-domain target tasks.
    \item To further highlight ICP as an effective proxy for gradient-based data attribution, we use both methods for dataset curation, and show that fine-tuning models on data highly-ranked by either method leads to similar performance.
    \item We explore the relationship between ICP and gradient-based attribution using synthetic data in a controlled setting. Our results show that the connection between these two methods can be isolated in toy settings, making it a potential path to bridge their gaps.
\end{enumerate}

%% file: sections/2_related_work.tex
\section{Related Work}\label{sec:related-work}

Obtaining high-quality training data is important for efficient model training \citep{lee2022deduplicating, sorscher2023neural, ye2024data, albalak2024survey}. One class of data attribution methods is gradient-based methods, such as influence functions \citep{koh2017understanding}, which utilize model gradients that estimate the influence of a training sample on model predictions. Despite being computationally expensive in LLM settings \citep{grosse2023studying}, gradient-based methods are effective for curating subsets of high-quality training data \citep{pruthi2020estimating, park2023trak, han-etal-2023-understanding, xia2024less, engstrom2024dsdm}.

Based on the phenomenon of transformers having in-context learning capabilities \citep{min-etal-2022-rethinking, han-etal-2023-understanding, bhattamishra2023understanding, liu2024understandingincontextcontrastive}, recent works have used ICP for training data selection \citep{rubin2022learning, nguyen2023incontext, iter-etal-2023-context, wettig2024qurating}. These methods involve measuring the model output likelihoods of the task given an in-context train sample, or prompting an LLM with questions to identity high-quality training data. For example, \citet{li2024shot} demonstrated that training on subsets of high-quality data using ICP leads to better performance than training on the entire dataset.

Since both gradient-based data attribution methods and ICP can be used effectively for data selection, a key component to connecting these ideas lies in a recent body of work which suggests that in-context learning implicitly performs gradient descent by constructing meta-gradients \citep{irie2022dualformneuralnetworks, dai-etal-2023-gpt, von2023transformers}. Specifically, these studies highlight the duality between a forward pass through a transformer attention head and linear layers trained by gradient descent, but rely on major assumptions; including linear attention, and limited analysis on this phenomena on MLP layers. Despite these assumptions, transformer outputs for synthetic in-context tasks, such as linear regression, mirror the predictions of algorithms that implement gradient descent \citep{akyürek2023learningalgorithm, garg2023transformersfunction, mahankali2023stepgradient}, making the relationship between ICP and gradient descent an open research area.

%% file: sections/3_background.tex
\section{Preliminaries}\label{sec:background}


In order to draw a connection between ICP and gradient-based data attribution, we first present three methods for data selection: influence functions \citep{koh2017understanding}, local datamodeling \citep{iter-etal-2023-context, yu2024mates} and ICP scoring \citep{li2024shot}. We begin by defining some notation: let  $\mathcal{D}_{train} = \{z_i\}_{i=1}^{N}$ be a set of training samples, where a sample $z_i = (x_i, y_i)$ contains an input and output. Similarly, let $\mathcal{D}_{test}= \{z'_i\}_{j=1}^{M}$ be a set of test samples. \\

\noindent \textbf{Method 1: Influence Functions} ~\citep{koh2017understanding} approximate changes in model predictions when samples are added/removed from the model's training data. To measure the influence of train sample $z \in \mathcal{D}_{train}$, the change in model parameters $\theta^*$ is approximated when $z$ is up-weighting by a small value $\epsilon$. Thus, the empirical risk minimization is:
\begin{align}
    \theta^*(\epsilon) = \argmin_{\theta} \frac{1}{N} \sum_{i=1}^{N} \mathcal{L} (z_i; \theta) + \epsilon \mathcal{L} (z; \theta),
\end{align}

\noindent which is also called the response function. We wish to find the change in parameters $\Delta \theta = \theta^*(\epsilon) - \theta^*$, which can be done via a first-order Taylor approximation to the response function at $\epsilon = 0$, which yields $\theta^*(\epsilon) - \theta^* \approx \epsilon \od{\theta^*(\epsilon)}{\epsilon} \bigg\rvert_{\epsilon = 0}$. Moreover, using the Implicit Function theorem, we get the influence of $z$ on $\theta^*$.
\begin{equation}
    I_{\theta^*} (z)  
    = \od{\theta^*(\epsilon)}{\epsilon} \bigg\rvert_{\epsilon = 0} 
    = -H^{-1} \nabla_{\theta} \mathcal{L} (z; \theta^*) ,
\end{equation}
\noindent where $H = \frac{1}{N} \sum_{i=1}^{N} \nabla_{\theta}^2  \mathcal{L} (z_i; \theta^*)$ and $z_i \in \mathcal{D}_{train}$. To quantify the influence of $z$ specifically on $z'$, we can measure influence with respect to $\mathcal{L}(z'; \theta)$, the loss on $z'$, which via the chain rule results in:
\begin{equation}\label{eqn:if}
    \text{Infl}(z', z) = \nabla \mathcal{L}(z'; \theta) H^{-1}  \nabla \mathcal{L}(z; \theta).
\end{equation}

\noindent Computing $H^{-1}$ expensive and unstable in non-convex loss function settings, such as for large deep learning models \citep{basu2021influence}. A simpler and more cost effective alternative \citep{pruthi2020estimating, xia2024less} is to drop the Hessian and only keep the \textit{inner product}:
\begin{equation}
    \text{Infl}_{\text{IP}}(z', z) =  \nabla \mathcal{L}(z'; \theta) \cdot \nabla \mathcal{L}(z; \theta).
\end{equation}

\noindent In particular, \cite{yang2024revisit} showed that despite dropping the Hessian, $\text{Infl}_{\text{IP}}$ exhibits good order-consistency with Infl. \\

\noindent \textbf{Method 2: Local Data Influence.} The influence of training sample $z \in \mathcal{D}_{train}$ towards a test sample $z'= (x', y')$ can also be measured using a one-step training score \citep{pruthi2020estimating, iter-etal-2023-context, yu2024mates}. Formally, this score is defined as:
\begin{equation}\label{eqn:ft}
    \text{Infl}_{\text{Loc}}(z, z') =  s_{zs}(z'; \hat{\theta}) - s_{zs}(z'; \theta). 
\end{equation}

\noindent where $\hat{\theta} = \theta -  \eta \nabla \mathcal{L}(\theta, z)$ denotes the parameters of a model trained on $z$ for single step with learning rate $\eta$. We denote $\text{s}_{\text{zs}}(z'; \theta) = \log p(y' | x'; \theta)$  as the zero-shot score (i.e., the model likelihood for the test sample output). Alternatively, the contribution of train sample $z$ towards an entire test set $\mathcal{D}_{test}$ can be aggregated as:

\small
\begin{equation}
    \text{Infl}_{\text{Loc}}(z, \mathcal{D}_{test}) 
    = \frac{1}{M} \sum_{j=1}^{M} \mathbbm{1}[s_{zs}(z'_{j}; \hat{\theta}) > s_{zs}(z'_{j}; \theta)].
\end{equation}
\normalsize

\noindent $\text{Infl}_{\text{Loc}}$ performs \textit{local} datamodeling since the influence of $z$ is measured by a single training step on an existing pre-trained model, rather than fully re-training the model with $z$. \\


\noindent \textbf{Method 3: In-Context Probing Score.} Leveraging the in-context learning abilities of LLMs, the importance of training sample $z$ can also be measured using a one-shot quality score introduced in \citet{li2024shot}. Formally, the ICP score is: 

\small
\begin{equation}\label{eqn:icp}
    \text{ICP}(z, \mathcal{D}_{test})
    = \frac{1}{M} \sum_{j=1}^{M} \mathbbm{1}[s_{os}(z'_{j} | z; \theta) > s_{zs}(z'_{j}; \theta)],
\end{equation}
\normalsize

\noindent where for a test sample $z' = (x', y')$, the one-shot score is defined as $\text{s}_{\text{os}}(z', z; \theta) = \log p(y'|z, x'; \theta)$, which is the model likelihood for the output of test sample $z'$ with $z$ as an in-context demonstration. \\

\noindent \textbf{Connecting ICP, $\text{Infl}_\text{Loc}$, and $\text{Infl}_\text{IP}$}. While all three methods can be used to measure the importance of training samples for a test task, they differ in computational efficiency. Notably, ICP is convenient since, unlike $\text{Infl}_\text{Loc}$, it requires no training, and, unlike $\text{Infl}_\text{IP}$, it does not access model gradients. Given the advantages of using ICP, we note the connection between ICP and $\text{Infl}_\text{IP}$ through $\text{Infl}_\text{Loc}$, which shows how ICP can be an efficient proxy for gradient-based data attribution.

First, we draw a connection between ICP and $\text{Infl}_\text{Loc}$ from recent works \citep{irie2022dualformneuralnetworks, dai-etal-2023-gpt, von2023transformers} which show that a linear attention head performs an implicit gradient descent update on in-context demonstrations. We present this construction below (details in Appendix \ref{sec:appendix_icp_implicit}):


\small
\begin{equation}
    Attn(K, V, q) \approx (W_{z'} + \Delta  W_{z}) q,
\end{equation}
\normalsize

\noindent where $W_{z'}$ represents the attention head weights for a test query $z'$. Notably, $\Delta W_{z}$ is the update for an in-context demonstration $z$, which is applied to attention head weights $W_{z'}$. Given the update of $z$ onto $W_{z'}$, this construction shares similarities with performing an actual gradient descent update of $z$ onto the model parameters. A resulting hypothesis is that for a model parameterized by $\theta$, we have: 

\begin{equation}\label{eqn:hypo}
    s_{os}(z' | z; \theta) \propto s_{zs}(z'; \hat{\theta}),
\end{equation}

\noindent where $\hat{\theta} = \theta -  \eta \nabla \mathcal{L}(z; \theta)$. That is, taking a training step on $z$ has similar effects on the model output likelihoods for $z'$ as using it as an in-context demonstration. Thus, if equation \ref{eqn:hypo} holds, then we have $\text{ICP}(z', z) \propto \text{Infl}_\text{Loc}(z', z)$. Moreover, connecting ICP and $\text{Infl}_{\text{IP}}$ is straightforward since $\text{Infl}_\text{IP}(z', z)$ is an approximation of $\text{Infl}_\text{Loc}(z', z)$, as noted in \citet{pruthi2020estimating}. As a result, we have:
\begin{equation}\label{eqn:icp-loc-infl}
\text{ICP}(z', z) \propto \text{Infl}_{\text{Loc}}(z', z) \approx \text{Infl}_{\text{IP}}(z', z).    
\end{equation}

\noindent The full derivation of this result is in Appendix \ref{sec:appendix-proofs}. An implication of equation \ref{eqn:icp-loc-infl} is that we expect positive correlation between ICP and $\text{Infl}_{\text{IP}}$ scores with respect to how they rank training data for test samples. In the next sections, we empirically explore this correlation.

%% file: sections/4_setup.tex
\section{Experiments on NLP Datasets}\label{sec:setup}\label{sec:experiments}

Given the connection between ICP and $\text{Infl}_{\text{IP}}$, we describe our experimental setup to analyze how well these two methods correlate in their rankings of influential training data. As noted in the previous section, a key component in connecting $\text{ICP}(z', z)$ with $\text{Infl}_{\text{IP}}(z', z)$ is the hypothesis that ICP performs a process akin to a gradient descent step on a train sample $z$ (i.e., $\text{ICP}(z, z') \propto \text{Infl}_{\text{Loc}}(z', z)$). A key question is whether this process occurs for any arbitrary $z$ and $z'$, since this would affect the correlation between $\text{ICP}(z', z)$ and $\text{Infl}_{\text{IP}}(z', z)$ rankings. To investigate this empirically, we vary $z$ and $z'$ to be ``in-domain" and ``out-of-domain", and examine the correlation step-by-step between $\text{ICP}$, $\text{Infl}_{\text{Loc}}$, and $\text{Infl}_{\text{IP}}$. Although ``in-domain" is loosely defined in NLP, two features that commonly define whether $z$ is in the same domain as $z'$ involve \textit{task} and \textit{content} similarity \citep{ramponi-plank-2020-neural}. 


Formally, we define a set of tasks $\{t_i\}_{i=1}^{T} \in \mathcal{T}$. Each task maps an input $x \in \mathcal{X}$ into a output $y \in \mathcal{Y}$ (i.e., $t: \mathcal{X} \rightarrow \mathcal{Y}$) to create a data sample $z = (x, y)$. We consider a set of train samples for task $t$, which we denote as $\mathcal{D}_{train}^{t}$, and a set of test samples for task $t'$, which we denote as $\mathcal{D}_{test}^{t'}$. We are interested in how the correlation between ICP and $\text{Infl}_{\text{IP}}$ changes as the following features vary between $\mathcal{D}_{train}^{t}$ and $\mathcal{D}_{test}^{t'}$:

\begin{enumerate}
    \item \textbf{Task Similarity}: as the train task $t$ and test task $t'$ change. In our experiments, we heuristically define the train and test tasks to be standard NLP tasks. In particular, we fix the test task be to instruction-following, and vary the train tasks to be instruction-following, QA/DocQA, and pretrain tasks, which we describe in detail in Section \ref{sec:exp-task-definitions}.
    
    \item \textbf{Content Similarity}: as the semantic similarity between train sample $z$ and test sample $z'$ change. In our experiments, we fix the train task $t$ and test task $t'$ to be the same, and vary the content using BertScore \citep{zhang2020bertscore}, a popular evaluation metric which measures similarity between two sequences using pretrained BERT embeddings. 
\end{enumerate}

\noindent Next, given $\mathcal{D}_{train}^{t}$ and $\mathcal{D}_{test}^{t'}$, we measure the Spearman correlation between ICP and $\text{Infl}_{\text{IP}}$ rankings. First, we obtain ICP and $\text{Infl}_{\text{IP}}$ scores of the all samples in $\mathcal{D}_{train}^{t}$ for test set $\mathcal{D}_{test}^{t'}$, which we denote as:
\begin{align}
    &\mathcal{S}_{\text{ICP}}(\mathcal{D}_{train}^{t}, \mathcal{D}_{test}^{t'}) \notag \\
    &\quad = \{\text{ICP}(z_i, \mathcal{D}_{test}^{t'}) | z_i \in \mathcal{D}_{train}^{t} \}_{i=1}^{N}, \\
    &\mathcal{S}_{\text{Infl}_{\text{IP}}}(\mathcal{D}_{train}^{t}, \mathcal{D}_{test}^{t'}) \notag \\
    &\quad = \{\text{Infl}_{\text{IP}}(z_i, \mathcal{D}_{test}^{t'})  | z_i \in \mathcal{D}_{train}^{t} \}_{i=1}^{N}.
\end{align}
\noindent Finally, we calculate the Spearman correlation between the ICP and $\text{Infl}_{\text{IP}}$ scores:
\small{
\begin{equation}
    \text{Spearman}(\mathcal{S}_{\text{ICP}}(\mathcal{D}_{\text{train}}^{t}, \mathcal{D}_{test}^{t'}), \mathcal{S}_{\text{Infl}_{\text{IP}}}(\mathcal{D}_{\text{train}}^{t}, \mathcal{D}_{\text{test}}^{t'})).
\end{equation}}
\normalsize

%% file: sections/5_experiments.tex
\input{tables/task-similarity}
\input{figures/content-similarity.tex}
\input{figures/task-similarity.tex}
\subsection{Datasets and Models}\label{sec:exp-task-definitions} 

In this section, we define a set of NLP tasks used in our experiments, which differ in objective and structure. We describe the datasets used for each task (see Table \ref{tab:task-examples} in Appendix \ref{sec:appendix-tables-figures} for examples), and also describe our models. \\

\noindent \textbf{Instruction Tasks}: Instruction-following requires a language model to generate an appropriate response by following an instruction (e.g., "Write a poem about the Autumn"), making it a key component in LLM research and real-world applications \citep{ouyang2022traininglanguagemodelsfollow, zhang2024instructiontuning}. For instruction tasks, we use the Alpaca dataset \citep{taori2023stanford}, which contains ~52K instruction demonstrations generated by GPT-4 following the Self-Instruct method \citep{wang2023selfinstruct}. \\


\noindent \textbf{QA/DocQA Tasks}: QA tasks are simple question-answering tasks without any context in the input (e.g., "What is the capital city of the U.S?"). They differ from instruction tasks since they may not explicitly provide an instruction in the task. DocQA tasks are question-answering tasks with additional context in the input (e.g., "Read following movie review and rate it: ..."). We sourced QA and DocQA tasks from PromptSource \citep{bach2022promptsource} dataset, which contains human-written prompts. We split the dataset into QA and DocQA datasets, using 9K and 8K examples, respectively. \\

\noindent \textbf{Pretrain Tasks:} Unlike the previous tasks, pretrain data is unstructured and does not contain any explicit questions or instructions. We sourced pretrain data from Minipile \citep{kaddour2023minipile}, which is a subset of the Pile \citep{gao2020pile} dataset curated for data diversity. We split the Minipile dataset into sequences of 256 tokens, and take a subset of 25K pretrain sequences.\\

\noindent \textbf{Train and Test Set Splits:} For each task, we use each dataset as the train set and take 100 samples from each dataset to form their respective test sets. \\ 

\noindent \textbf{Models:} Across all experiments, we calculate ICP, $\text{Infl}_{\text{Loc}}$, and $\text{Infl}_{\text{IP}}$ scores using Pythia-1b-deduped \cite{biderman2023pythia} and Llama-3.2-3B. For calculating $\text{Infl}_{\text{Loc}}$, we set the learning rate to 2e-5.

\subsection{Task Similarity Results}\label{sec:exp-task}

Table \ref{tab:task-similarity} shows that ICP/$\text{Infl}_{\text{IP}}$ correlation is high when the task types of the train and test datasets are the same. This correlation decreases significantly when the task type of the train and test datasets differ. Note that $\text{Infl}_{\text{Loc}}$/$\text{Infl}_{\text{IP}}$ correlation remains high overall regardless of task difference between the train and test datasets. This suggests that $\text{Infl}_{\text{Loc}}$ is a close approximation for $\text{Infl}_{\text{IP}}$, and that the breaking point lies between ICP and $\text{Infl}_{\text{Loc}}$. Thus, the hypothesis introduced in section \ref{sec:background} that ICP performs a gradient descent-like step does not hold when the train and test task types differ.

\input{figures/alpaca-icp-if-ft-corr}
\input{tables/winrate-table-equal-samples}

\subsection{Content Similarity Results}\label{sec:exp-content}

Figure \ref{fig:content-similarity} shows that ICP/$\text{Infl}_{\text{IP}}$ correlation decreases as the content similarity (i.e., BertScore) decreases between the train and test samples. Moreoever, $\text{Infl}_{\text{Loc}}$/$\text{Infl}_{\text{IP}}$ correlation remains high overall, and does not change much as content similarity decreases, which implies that that $\text{Infl}_{\text{Loc}}$ is a robust approximation for $\text{Infl}_{\text{IP}}$ regardless of content similarity. Similar to task type, the breaking point again lies between ICP and $\text{Infl}_{\text{Loc}}$ when the content similarity between the train and test samples differ.

\subsection{Task vs. Content Similarity Results}
Since both task type and content affects ICP/$\text{Infl}_{\text{IP}}$ correlation, we vary both features simultaneously and observe its impact. In Figure \ref{fig:task-similarity}, we examine ICP/$\text{Infl}_{\text{IP}}$ correlation as the BertScore decreases between the Alpaca test samples and train samples from the other previously defined tasks. Figure \ref{fig:task-similarity} shows no increase in ICP/$\text{Infl}_{\text{IP}}$ correlation as BertScore increases. Therefore, if the test and train tasks are different, then increasing content similarity does not result in better ICP/$\text{Infl}_{\text{IP}}$ correlation.

%% file: tables/task-similarity.tex
\begin{table*}
\caption{Spearman correlation between ICP, $\text{Infl}_{\text{Loc}}$, and $\text{Infl}_{\text{IP}}$ using test samples from the Alpaca dataset and train samples from Alpaca, UltraChat, QA, DocQA, and pretrain datasets. All p-values are $<.05$.}
\label{tab:task-similarity}
\resizebox{\textwidth}{!}{
\begin{tabular}{lcccccccc}
    \toprule
    \multicolumn{1}{c}{} &\multicolumn{2}{c}{\textbf{Alpaca}} & \multicolumn{2}{c}{\textbf{QA}} & \multicolumn{2}{c}{\textbf{DocQA}} & \multicolumn{2}{c}{\textbf{Pretrain}} \\
    \cmidrule(lr){2-3}\cmidrule(lr){4-5}\cmidrule(lr){6-7}\cmidrule(lr){8-9}
    & Pythia-1b & Llama-3.2-3B & Pythia-1b & Llama-3.2-3B & Pythia-1b & Llama-3.2-3B & Pythia-1b & Llama-3.2-3B \\
    \midrule
    ICP/$\text{Infl}_{\text{IP}}$ & 0.73 & 0.54 & 0.10 & 0.10 & 0.21 & 0.13 & 0.07 & 0.12 \\
    ICP/$\text{Infl}_{\text{Loc}}$ & 0.61 & 0.36 & 0.10 & 0.13 & 0.17 & 0.26 & 0.03 & 0.05 \\
    $\text{Infl}_{\text{Loc}}$/$\text{Infl}_{\text{IP}}$ & 0.78 & 0.57 & 0.78 & 0.87 & 0.86 & 0.88 & 0.65 & 0.63 \\
    \bottomrule
\end{tabular}
}
\end{table*}

%% file: figures/content-similarity.tex
\begin{figure*}[t!]
\nocaption
  \begin{subfigure}{0.24\linewidth}
    \includegraphics[width=\linewidth]{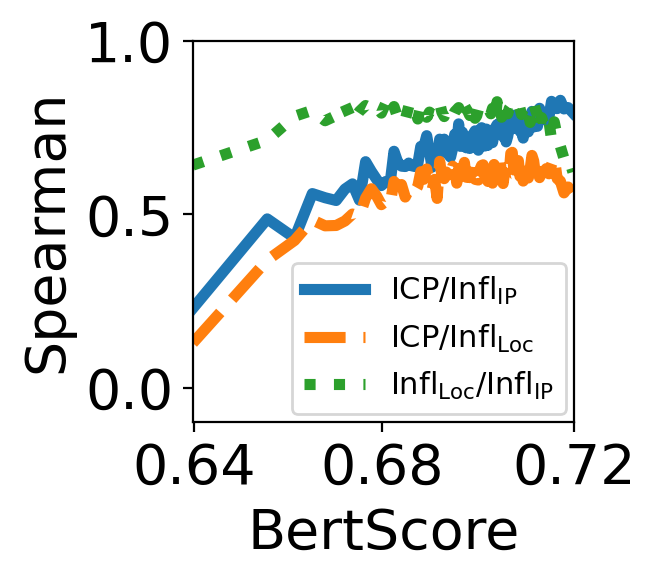}
    \caption{Alpaca (Pythia)}
    \label{fig:content-similarity-alpaca-pythia}
  \end{subfigure}
  \begin{subfigure}{0.24\linewidth}
    \includegraphics[width=\linewidth]{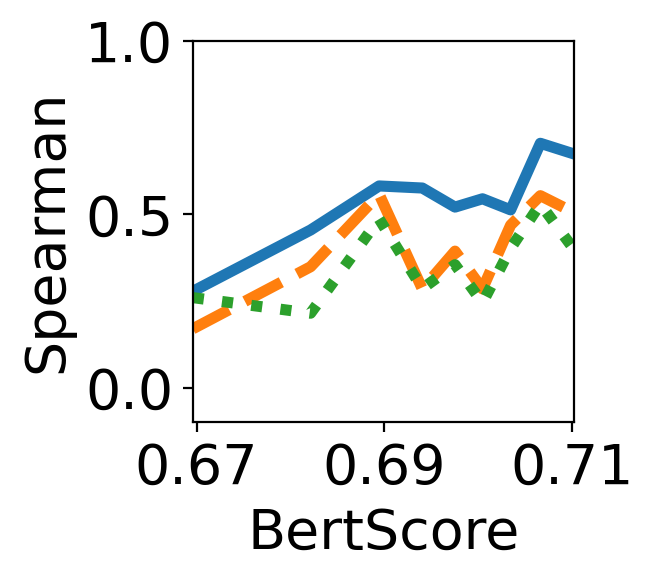}
    \caption{Alpaca (Llama)}
    \label{fig:content-similarity-alpaca-llama}
  \end{subfigure}
  \begin{subfigure}{0.24\linewidth}
    \includegraphics[width=\linewidth]{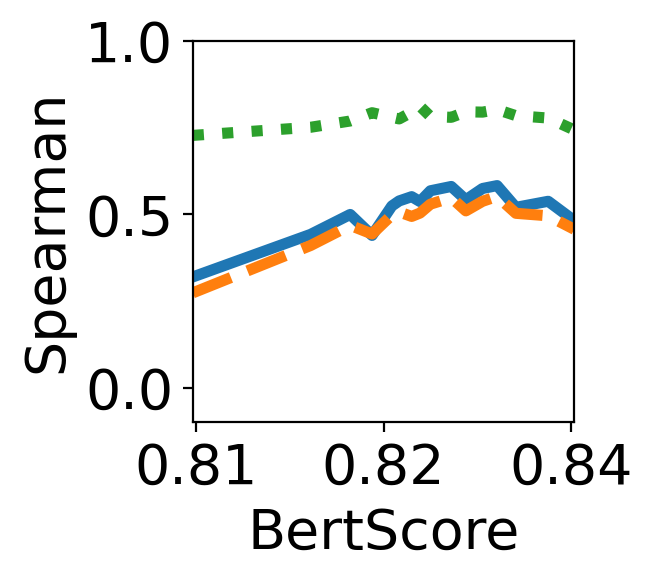}
    \caption{QA (Pythia)}
    \label{fig:content-similarity-qa-pythia}
  \end{subfigure}
  \begin{subfigure}{0.24\linewidth}
    \includegraphics[width=\linewidth]{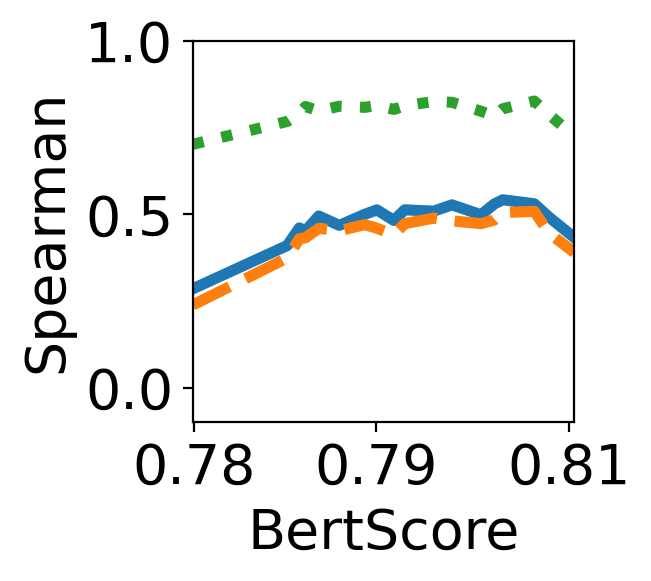}
    \caption{DocQA (Pythia)}
    \label{fig:content-similarity-docqa-pythia}
  \end{subfigure}
  \caption{Correlation analysis between ICP, $\text{Infl}_{\text{Loc}}$, and $\text{Infl}_{\text{IP}}$ (aggregated across groups of 500 samples) with respect to content similarity (BertScore) using test and train samples from the same task.}
  \label{fig:content-similarity}
\end{figure*}

%% file: figures/task-similarity.tex
\begin{figure*}[t!]
\nocaption
   \begin{subfigure}{0.24\linewidth}
    \includegraphics[width=\linewidth]{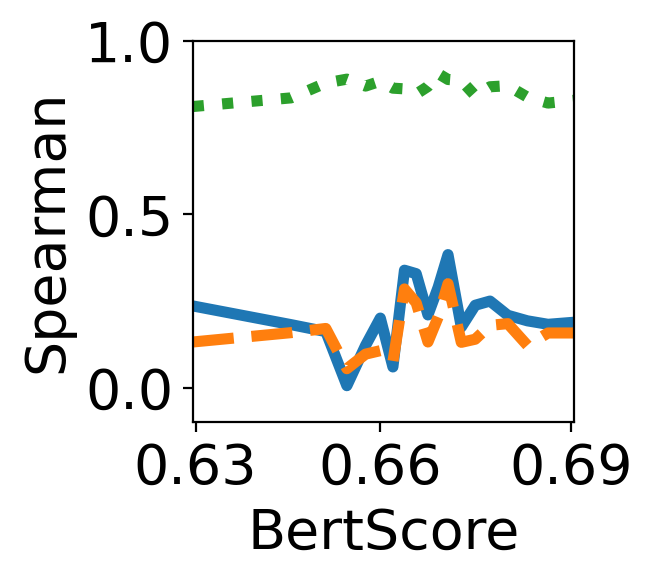}
    \caption{DocQA (Pythia)}
    \label{fig:bertscore_kmeans_doc_qa_pythia}
  \end{subfigure}
  \begin{subfigure}{0.24\linewidth}
    \includegraphics[width=\linewidth]{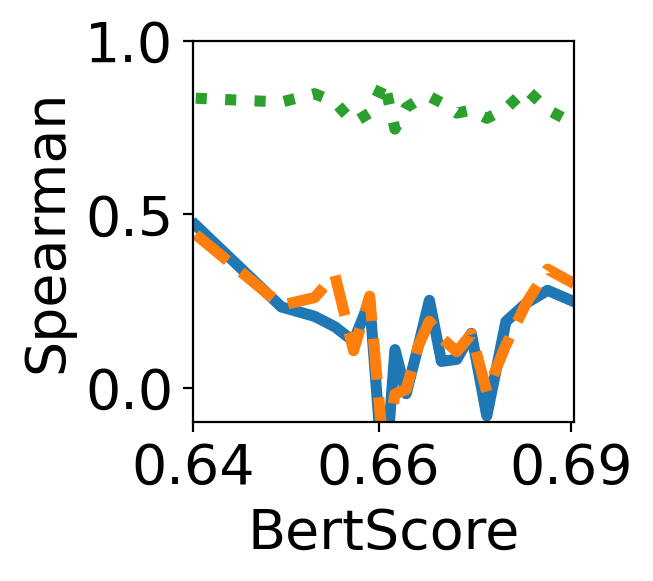}
    \caption{DocQA (Llama)}
    \label{fig:bertscore_kmeans_doc_qa_llama}
  \end{subfigure}
  \begin{subfigure}{0.24\linewidth}
    \includegraphics[width=\linewidth]{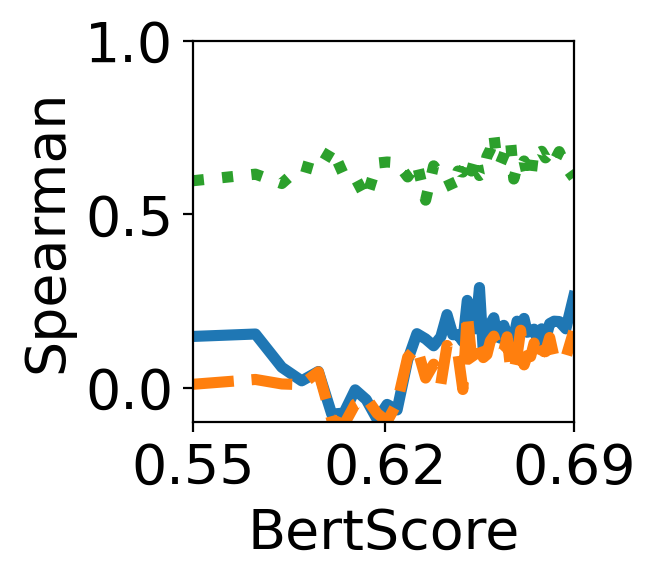}
    \caption{Pretrain (Pythia)}
    \label{fig:bertscore_kmeans_minipile_pythia}
  \end{subfigure}
  \begin{subfigure}{0.24\linewidth}
    \includegraphics[width=\linewidth]{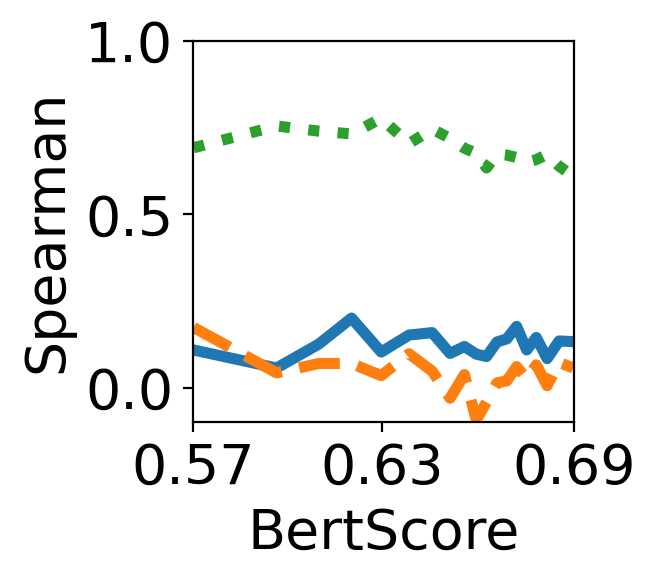}
    \caption{Pretrain (Llama)}
    \label{fig:bertscore_kmeans_minipile_llama}
  \end{subfigure}
  \caption{Correlation analysis between ICP, $\text{Infl}_{\text{Loc}}$, and $\text{Infl}_{\text{IP}}$ (aggregated across groups of 500 samples) with respect to content similarity (BertScore) using test samples from Alpaca and training samples from DocQA/Pretrain datasets. Additional analysis in Appendix \ref{sec:appendix-tables-figures}.}
  \label{fig:task-similarity}
\end{figure*}

%% file: figures/alpaca-icp-if-ft-corr.tex
\begin{figure*}[t!]
\nocaption
\centering 
  \begin{subfigure}{0.24\linewidth}
    \centering 
    \includegraphics[width=\linewidth]{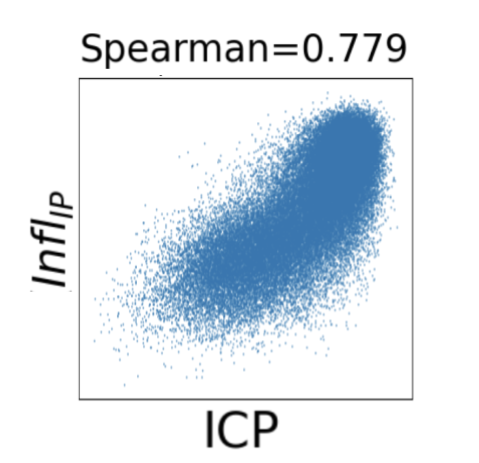}
    \caption{ICP/$\text{Infl}_{\text{IP}}$ scores.}
    \label{fig:correlation-graph}
  \end{subfigure}
  \begin{subfigure}{0.24\linewidth}
    \includegraphics[width=\linewidth]{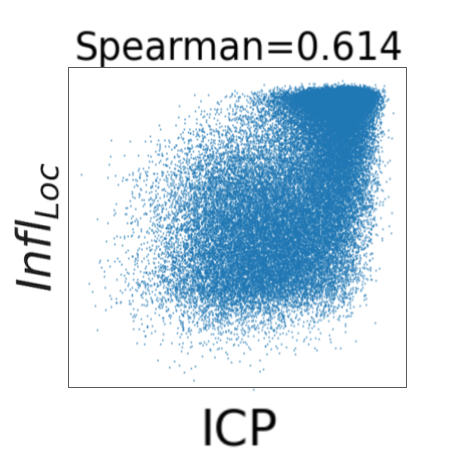}
    \caption{ICP/$\text{Infl}_{\text{Loc}}$ scores.}
    \label{fig:icp-ft-correlation} 
  \end{subfigure}
  \begin{subfigure}{0.24\linewidth}
    \centering 
    \includegraphics[width=\linewidth]{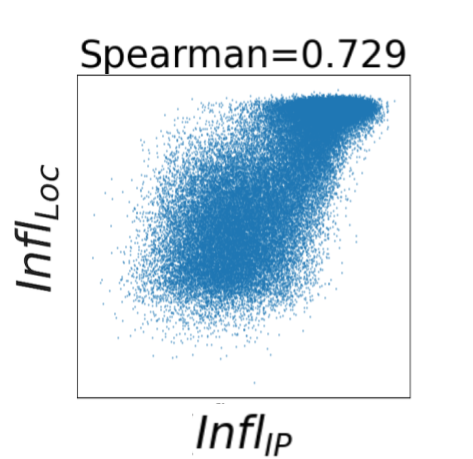}
    \caption{$\text{Infl}_{\text{IP}}$/$\text{Infl}_{\text{Loc}}$ scores.}
    \label{fig:infl-ft-corrrelation}
  \end{subfigure}
  \begin{subfigure}{0.22\linewidth}
    \centering 
    \includegraphics[width=\linewidth]{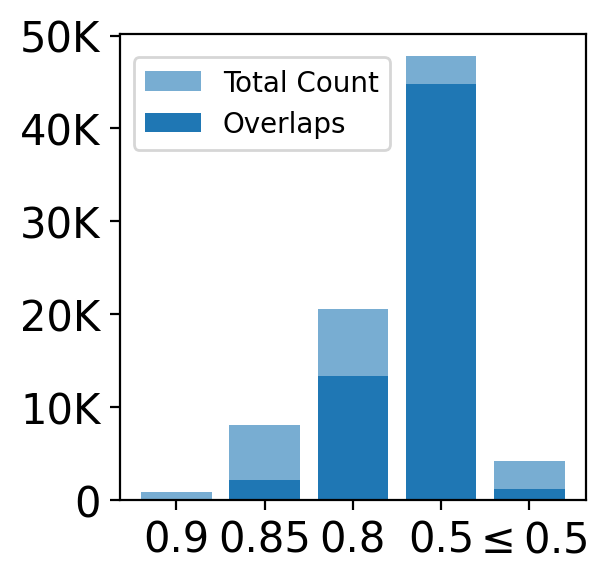}
    \caption{ICP/$\text{Infl}_{\text{IP}}$ overlaps.}
    \label{fig:overlap-graph}
  \end{subfigure}
  \caption{Correlation analysis between rankings on the instructions from the Alpaca dataset assigned by ICP, $\text{Infl}_{\text{Loc}}$, and  $\text{Infl}_{\text{IP}}$. All p-values are $<.05$.}
  \label{fig:alpaca-icp-if-corr}
\end{figure*}


%% file: tables/winrate-table-equal-samples.tex
\begin{table*}[t!]
\centering
\caption{Results (winrates) evaluated on the Alpaca Eval dataset after being finetuned on data selected by ICP and $\text{Infl}_\text{IP}$. The highest winrate in each column is marked with $*$ for ICP and $\dagger$ for $\text{Infl}_\text{IP}$.}
\scalebox{0.95}{
\begin{tabular}{l|r|llllll}
    \toprule
    \textbf{Score Bin} & \textbf{Method} & \textbf{Helpful Base} & \textbf{Koala} & \textbf{Self Instruct} & \textbf{Oasst} & \textbf{Vicunna} & \textbf{Overall} \\
    \midrule
     $\leq 0.5$ & ICP & $54.26$ & $62.99$ & $56.04$ & $56.15$ & $51.25$ & $56.65$ \\
                & $\text{Infl}_\text{IP}$ & $51.56$ & $62.18$ & $54.76$ & $54.55$ & $43.75$ & $54.54$ \\
     \hline
     $> 0.5$  & ICP & $57.03$ & $62.99$ & $\mathbf{61.35}$ & $63.10$ & $68.75$ & $62.12$ \\
              & $\text{Infl}_\text{IP}$ & $60.94$ & $65.16$ & $\mathbf{65.08\dagger}$ & $60.42$ & $52.50$ & $62.09$ \\
     \hline
     $> 0.8$  & ICP & $61.42$ & $59.35$ & $58.73$ & $\mathbf{64.17^*}$ & $67.50$ & $61.42$ \\
              & $\text{Infl}_\text{IP}$ & $62.79$ & $\mathbf{68.18\dagger}$ & $62.3$ & $61.70$ & $\mathbf{68.75\dagger}$ & $\mathbf{64.02\dagger}$\\
     \hline
     $> 0.85$ & ICP & $\mathbf{62.79^*}$ & $\mathbf{62.58^*}$ & $60.16$ & $61.17$ & $\mathbf{70.00^*}$ & $\mathbf{62.26^*}$ \\
              & $\text{Infl}_\text{IP}$ & $\mathbf{65.12\dagger}$ & $67.95$ & $56.35$ & $\mathbf{65.42\dagger}$ & $65.00$ & $62.98$ \\
     \hline
     $> 0.9$ & ICP & $52.34$ & $60.13$ & $49.79$ & $49.46$ & $46.25$ & $51.77$ \\
             & $\text{Infl}_\text{IP}$ & $61.24$ & $57.14$ & $53.60$ & $55.08$ & $55.00$ & $56.00$\\
     \bottomrule
\end{tabular}
}
\label{tab:winrates}
\end{table*}


%% file: sections/6_data_selection_experiments.tex
\section{ICP for Data Selection}\label{sec:in-domain}

In the previous section, we showed that ICP and $\text{Infl}_{\text{IP}}$ correlate well in how they rank influential training data when the train set shares the same task type as the test set. This is is strongly reflected in the Alpaca instruction-following dataset (see Table \ref{tab:task-similarity} and Figure \ref{fig:alpaca-icp-if-corr}). In this case, it is possible that ICP can serve as a proxy for $\text{Infl}_{\text{IP}}$. This has promising implications: compute costs for ICP is significantly cheaper than $\text{Infl}_{\text{IP}}$. For instance, to score the entire Alpaca dataset with Pythia-1b, ICP incurred a total of 10 GPU hours while $\text{Infl}_{\text{IP}}$ incurred 90 hours.

However, since ICP and $\text{Infl}_{\text{IP}}$ rankings are not entirely aligned, we further compare them by using both methods to curate datasets for instruction-tuning. Following the same setup as \citet{li2024shot}, we first we obtained ICP scores (and in our case, $\text{Infl}_{\text{IP}}$ scores as well) for all training samples in the Alpaca dataset using the K-Means-100 dataset (a subset of 100 diverse instructions from the Alpaca dataset created by \citet{li2024shot}) as the test set. We use the ICP and $\text{Infl}_{\text{IP}}$ scored training samples for instruction-tuning according the following procedure: \\


\noindent \textbf{Finetuning Datasets:} After obtaining ICP and $\text{Infl}_{\text{IP}}$ scores (reminder: ICP $\in [0,1]$) for the Alpaca dataset, we create ICP score bins of $\leq 0.5, >0.5, >0.8, >0.85, >0.9$. We used the number of samples in each score bin as threshold cutoffs for $\text{Infl}_{\text{IP}}$. For example, if the $>0.9$ ICP score bin had \textit{k} training samples, then we also treated the top \textit{k} samples from $\text{Infl}_{\text{IP}}$ as the equivalent bin. We treat all bins as separate datasets, and randomly sample 700 demonstrations from each dataset for fine-tuning. \\

\noindent\textbf{Training:} We use the Adam optimizer with a batch size of 64 and lr=2e-7 to fine-tune Pythia-1b-deduped for 3 epochs. This is done separately for ICP and $\text{Infl}_{\text{IP}}$ for each score bin. \\

\noindent\textbf{Evaluation}: We use the Alpaca Eval dataset \citep{li2023alpacaeval, dubois2024alpacafarm}, which has 805 instruction demonstrations (details in Appendix \ref{sec:appendix-tables-figures}). The evaluation metric for the Alpaca Eval dataset is winrate \citep{li2023alpacaeval}, which is the expected preference of a human (or LLM) annotator for a model's response compared to a baseline model's response. We follow the same setup as \citet{li2024shot}, and use GPT-4 Turbo as the annotator. Winrates are calculated by comparing our fine-tuned models to Pythia-1b-deduped. \\

\noindent \textbf{Results:} First, we note that Figure \ref{fig:overlap-graph} shows good overlap between instructions selected by both methods across different score bins, which suggests that ICP and $\text{Infl}_{\text{IP}}$ have high agreement on instruction quality and valuation. Next, our results in Table \ref{tab:winrates} shows that fine-tuning on instruction data selected by ICP and $\text{Infl}_{\text{IP}}$ result in similar model performance among different score ranking bins, and overall performance for ICP and $\text{Infl}_{\text{IP}}$ both peaked around similar score bins (i.e.,  $>0.8$ and $>0.85$). Examples of top-ranked instructions selected by ICP and $\text{Infl}_{\text{IP}}$ are shown in Table \ref{tab:instruction-examples} in Appendix \ref{sec:appendix-tables-figures}. Overall, our findings highlight the consistency between ICP and $\text{Infl}_{\text{IP}}$ in selecting high-quality instructions for when task type between the training data and target task are similar, which shows that bridging the gap between ICP and $\text{Infl}_{\text{IP}}$ has promising implications.


%% file: sections/7_synthetic_experiments.tex
\section{Synthetic Study}\label{sec:syn-experiments}

\input{figures/lin-func-input-task-corr}
While the results in Section \ref{sec:experiments} show that task and content similarity affects how well ICP approximates $\text{Infl}_{\text{IP}}$ for NLP tasks, in this section, we further study the correlation between ICP and $\text{Infl}_{\text{IP}}$ in a constrained and well-defined setting using linear regression tasks. Studying the correlation between ICP and $\text{Infl}_{\text{IP}}$ in this setting offers a significant advantage: unlike standard NLP tasks, we can easily isolate and control both the task and content similarity of a linear regression task, allowing for a more granular examination of the correlation.


In this setup, we first sample a function parameter $w \in \mathbb{R}^d$ and an input $x \in \mathbb{R}^d$ separately from an isotropic Gaussian distribution $\mathcal{N}(0, I_d)$. The output  $y = f(x) = w^T x$. Thus, the function parameter $w$ is the \textit{task} since it defines the relationship between the input and output. Given $k$ in-context demonstrations $\{(x_1, y_1), ..., (x_k, y_k)\} \in \mathcal{D}_{train}^w$ and a test sample $(x', y') \in \mathcal{D}_{test}^{w'}$, we prompt the model to predict $y' = f(x') = w^{'T} x'$ using input prompt sequence $(x_1, y_1..., x_k, y_k, x')$. Next, we describe our experiments where we isolate and vary \textit{task type}  and \textit{content} to examine how well ICP rankings correlate with $\text{Infl}_{\text{IP}}$ rankings.\\

\noindent \textbf{Task Similarity Experiment:} Given a test sample $(x',y')$ with function parameter $w'$, we randomly draw sets of $k$ in-context demonstrations $\{(x_1, y_1), ..., (x_k, y_k)\}$ where $y_i = w^T x_i$ and $cos\_sim(w, w') = c$ for $i = 1, ..., k$. We vary $c$ from 0 to 9 with increments of 0.1, and also set $c=0.99$ and $c=0.999$ to examine cases where the training inputs are very close to the test input. \\

\noindent\textbf{Content Similarity Experiment:} Given a test sample $(x',y')$ with function parameter $w'$, we randomly draw sets of $k$ in-context demonstrations $\{(x_1, y_1), ..., (x_k, y_k)\}$ where $y_i = {w'}^T x_i$ and $cos\_sim(x_i, x') = c$ for $i = 1, ..., k$. We test for the same values of $c$, and use the same dataset generation process as mentioned above.\\

\noindent\textbf{Dataset Generation:} For both task and content similarity, we generate datasets using the following process: we first create 10 test inputs and 10 test function parameters. Using each test input and parameter pairing, we generate 1200 prompt sequences with $k$ demonstrations with varying task or content similarity with respect to the test query. This is repeated for each $k \in \{1, 5, 10, 20, 40\}$. \\

\noindent \textbf{Model:} We use the model provided by \citet{garg2023transformersfunction}, a decoder-only Transformer architecture (9.5M parameters), which is pre-trained on linear function classes. The model was trained for 500k steps and batch size of 64, where prompts sequences were randomly sampled for each step. \\

\noindent \textbf{Evaluation:} Given a test sample $z' = (x', y')$ and train sample $z = (x, y)$, we evaluate the model output $\hat{y'}$ against $y'$ using mean squared error (MSE) loss. For the synthetic data experiments, we set $\text{ICP}(z', z) = \text{MSE}(\hat{y}'; \theta) - \text{MSE}(\hat{y}'| y; \theta)$, where $\theta$ denotes the model parameters. Similarity, for $\text{Infl}_{\text{IP}}$ we set the loss function to be MSE such that $\text{Infl}_{\text{IP}}(z', z) = \nabla_{\theta} MSE(\hat{y}'; \theta) \cdot \nabla_{\theta} MSE(\hat{y}; \theta)$. \\


\noindent \textbf{Results:} We observe the effects of task and content similarity and ICP/$\text{Infl}_{\text{IP}}$ correlation in Figure \ref{fig:lin-func-input-task-corr}. When the train and test function parameters (i.e., task) are the same, ICP/$\text{Infl}_{\text{IP}}$ correlation is high, given that the content similarity is not low (Fig. \ref{fig:lin_func_input_icp_infl_n1}). However, when the train and test tasks are different, ICP/$\text{Infl}_{\text{IP}}$ correlation is low (Fig. \ref{fig:lin_func_task_icp_infl_n1}). In the case where both task and content similarity are varied (Fig. \ref{fig:lin-func-both-corr}), having greater content similarity can offset task disparity between the train and test samples. In addition, we note that as the number of in-context demonstrations in the prompt sequences increases (see Fig. \ref{fig:lin-func-input-task-corr-all} in Appendix \ref{sec:appendix-tables-figures}), the connection between ICP and $\text{Infl}_{\text{IP}}$ breaks. This can be due to group effects, where ICP and $\text{Infl}_{\text{IP}}$ provide different rankings for a group of training samples versus individual training samples. 

Overall, our synthetic experiments show that ICP correlates well with $\text{Infl}_{\text{IP}}$ when the train and test samples share the same task (i.e., function parameter), which is similar to our observation for NLP tasks in Section \ref{sec:experiments}. Given that this trend appears in both standard NLP and synthetic data settings, this highlights that the relationship between ICP and $\text{Infl}_{\text{IP}}$ can be studied from different angles. For instance, future work can explore additional cases for when connection between ICP and $\text{Infl}_{\text{IP}}$ breaks using more complex function classes, from both theoretical and empirical perspectives.

\input{figures/lin-func-both-corr}


%% file: figures/lin-func-input-task-corr.tex
\begin{figure}[th!]
  \centering
  \nocaption
  \begin{subfigure}{0.48\linewidth}
    \includegraphics[width=\linewidth]{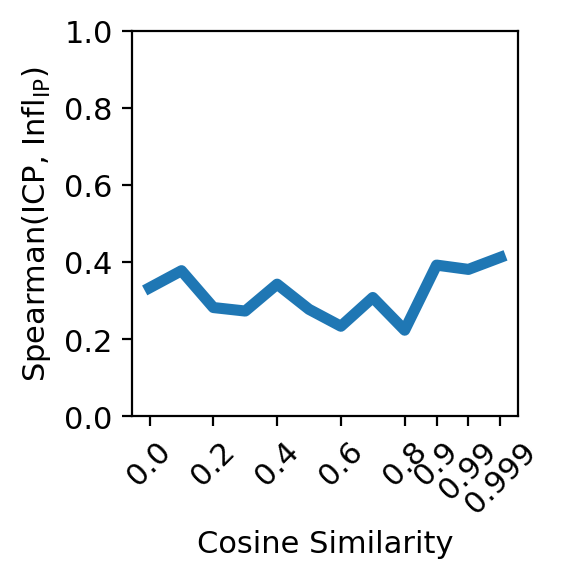}
    \caption{Task Similarity}
    \label{fig:lin_func_task_icp_infl_n1} 
  \end{subfigure}
  \begin{subfigure}{0.49\linewidth}
    \includegraphics[width=\linewidth]{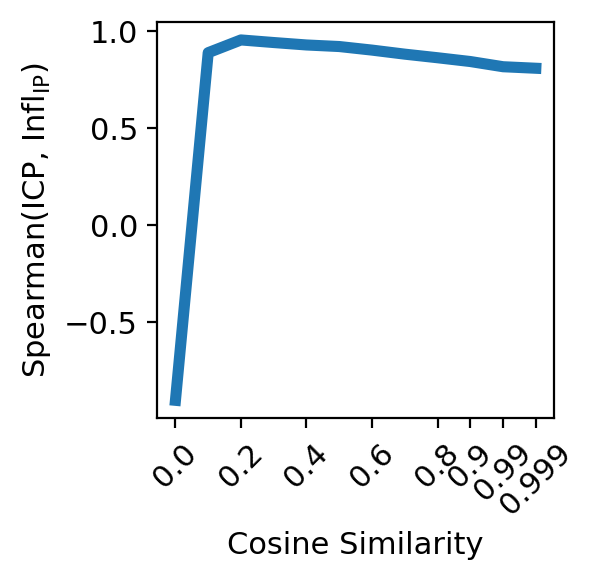}
    \caption{Content Similarity}
    \label{fig:lin_func_input_icp_infl_n1} 
  \end{subfigure}
  \caption{Correlation analysis between ICP and $\text{Infl}_{\text{IP}}$ as the task/content similarity of a single training demonstration vary with respect to the test query.}
  \label{fig:lin-func-input-task-corr}
\end{figure}

%% file: figures/lin-func-both-corr.tex
\begin{figure}[t!]
  \centering
  \nocaption
  \begin{subfigure}{\linewidth}
    \includegraphics[width=\linewidth]{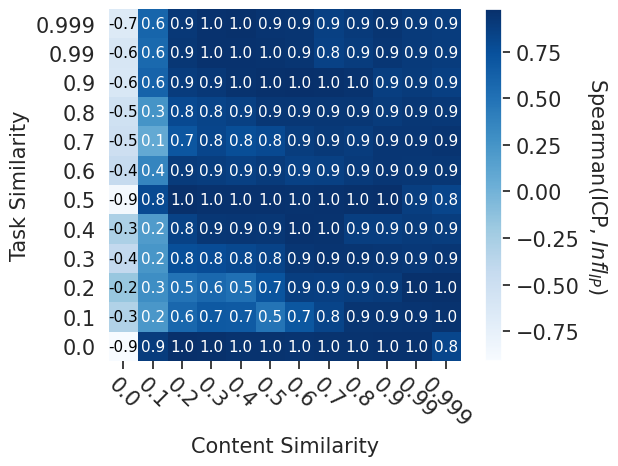}
  \end{subfigure}
  \caption{Correlation analysis between ICP and $\text{Infl}_{\text{IP}}$ as \textit{both} the task and content similarity of a single training demonstration vary with respect to the test query.}
  \label{fig:lin-func-both-corr}
\end{figure}

%% file: sections/8_conclusion.tex
\section{Conclusion}

In this paper we have examined the connection between ICP and gradient-based data attribution. Empirically, we have shown that ICP can serve as a proxy for influence functions in an in-domain data setting, where the train and target data have similar task type and content. As a result, this offers a possible explanation for why in-context probing is effective for data selection. We show that fine-tuning on influential data selected by both methods lead to similar downstream performance on instruction-following, which highlights a use case of ICP as a proxy for gradient-based data attribution. We furthermore explore their connection in a synthetic data setting, and observe similar results as the standard NLP data setting, paving the way for future work to explore this connection from theoretical angles. There are several lines of work that can further explore this phenomenon. For instance, finding methods to check whether ICP approximates gradient-based data attribution methods for black-box models. In addition, an important problem is how these two methods compare for selecting groups of training samples.

%% file: sections/ethics_limitations.tex
\section{Ethics and Limitations}
\label{sec:ethics_limitations}
First, we highlight limitations to our work. Our experiments are conducted using Pythia-1b deduped and LlaMa-3.2 3B. As model sizes change, the question of whether one data selection method triumphs over the other is an area for exploration. We also note our evaluation metric (winrate) for our instruction-tuning experiments rely on LLM annotation, and may be subject to LLM bias as mentioned in \citet{dubois2024lengthcontrolled}. Since our work involves understanding data valuation in language models, we note that language models themselves can be susceptible to biases. We hope that this work can lead to future work in understanding the mechanisms of LLMs. Further insight in that realm may be beneficial in understanding model predictions, especially when considering LLM safety, toxicity, and biases. 

%% file: sections/acknowledgements.tex
\section*{Acknowledgements}

We would like to thank Juhan Bae for providing insight on adapting influence function computations for large language models. We would also like to thank Jacob Springer for his insights.

%% file: sections/appendix.tex
\onecolumn

\section*{Appendix}

The appendix covers supporting information for our paper. In Section \ref{sec:appendix-proofs} we provide details for the connection between ICP and $\text{Infl}_{\text{IP}}$. In Section \ref{sec:appendix-tables-figures} provide all additional tables or figures referred to throughout the paper.

\input{sections/appendix_a}

\input{sections/appendix_b}
\input{sections/appendix_c}

%% file: sections/appendix_a.tex
\section{Connecting ICP, \texorpdfstring{$\text{Infl}_{\text{Loc}}$ }{INFL LOC} and   \texorpdfstring{$\text{Infl}_{\text{IP}}$}{INFL IP}}
\label{sec:appendix-proofs}

In this section, we provide the full details for how ICP connects to $\text{Infl}_{\text{IP}}$. As described in section \ref{sec:background}, for a train sample $z$ and test sample $z'$, we assume the the hypothesis $s_{os}(z' | z; \theta) \propto s_{zs}(z'; \hat{\theta})$ holds, where $\hat{\theta} - \theta - \eta \nabla \mathcal{L}(z; \theta)$. As a result of this hypothesis, we have:

\begin{equation}
\text{ICP}(z', z) \propto \text{Infl}_{\text{Loc}}(z', z) \approx \text{Infl}_{\text{IP}}(z', z)
\end{equation}

\noindent The first step is to show $\text{ICP}(z', z) \propto \text{Infl}_{\text{Loc}}(z', z)$. As noted in section \ref{sec:background}, given train sample $z$ and test sample $z'$, we assume that 
$s_{os}(z' | z; \theta) \propto s_{zs}(z'; \hat{\theta})$, where $\hat{\theta} - \theta - \eta \nabla \mathcal{L}(z; \theta)$. As a result, we have 

\begin{align}\label{eqn:icp-loc}
    \text{ICP}(z', z) 
    &= s_{os}(z'| z; \theta) - s_{zs}(z; \theta)  \notag \\
    &\propto s_{zs}(z'; \hat{\theta}) - s_{zs}(z'; \theta) \quad \text{by } s_{os}(z' | z; \theta) \propto s_{zs}(z'; \hat{\theta})  \notag \\
    &= \text{Infl}_{\text{Loc}}(z', z)
\end{align}

\noindent Next, to connect $\text{Infl}_{\text{Loc}}(z', z)$ with $\text{Infl}_{\text{IP}}(z', z)$, we begin by noting a derivation from \citet{pruthi2020estimating}:

\begin{lemma}\label{lemma:1}\citep{pruthi2020estimating}
Suppose we have a LLM with parameters $\theta$. We perform a gradient descent step with training sample $z$ with learning rate $\eta$ such that $\hat{\theta} = \theta -  \eta \nabla \mathcal{L}(z; \theta)$. Then,

\begin{equation}
    \mathcal{L}(z';\theta) - \mathcal{L}(z'; \hat{\theta}) \approx \nabla \mathcal{L}(z'; \theta) \cdot \nabla \mathcal{L}(z; \theta) 
\end{equation}

    
\noindent Proof: First, we consider the change in loss of $z'$ using a first-order approximation:
\begin{align}
    \mathcal{L}(z'; \hat{\theta})  
    &= \mathcal{L}(z'; \theta) + \nabla \mathcal{L}(z'; \theta) \dot (\hat{\theta} - \theta) + \mathcal{O}(|| \hat{\theta} - \theta||^2) \\
    \mathcal{L}(z'; \theta) - \mathcal{L}(z'; \hat{\theta})  
    &= - \nabla \mathcal{L}(z'; \theta) \dot (\hat{\theta} - \theta) + \mathcal{O}(||\hat{\theta} - \theta||^2)
\end{align}
\noindent Next, suppose a gradient descent step is taken on training sample $z$, and the model parameters are updated as: $\hat{\theta} = \theta -  \eta \nabla \mathcal{L}(z; \theta)$. Thus, we have $\hat{\theta} - \theta = -\eta \nabla \mathcal{L}(z; \theta)$, and the change in loss can be written as
\begin{align}
    \mathcal{L}(z'; \theta) - \mathcal{L}(z'; \hat{\theta})  
    &\approx \eta \nabla \mathcal{L}(z'; \theta) \cdot  \nabla \mathcal{L}(z; \theta) \propto \nabla \mathcal{L}(z'; \theta) \cdot  \nabla \mathcal{L}(z; \theta)
\end{align}

\noindent Given that $\eta$ is a constant. 
\end{lemma}

\noindent Next, to  connect $\text{Infl}_{\text{Loc}}(z', z)$ with $\text{Infl}_{\text{IP}}(z', z)$, we have: 

\begin{align}\label{eqn:loc-infl}
    \text{Infl}_{\text{Loc}}(z', z) \notag
    &=  s_{zs}(z'; \hat{\theta}) - s_{zs}(z'; \theta)  \notag \\
    &= \mathcal{L}(z'; \theta) - \mathcal{L}(z'; \hat{\theta})  \notag \\
    &\approx \nabla \mathcal{L}(z'; \theta) \cdot  \nabla \mathcal{L}(z; \theta)  \quad \text{by Lemma \ref{lemma:1}} \notag \\
    &= \text{Infl}_{\text{IP}}(z', z)
\end{align}

\noindent Finally, putting together equations \ref{eqn:icp-loc} and \ref{eqn:loc-infl}, we have $\text{ICP}(z', z) \propto \text{Infl}_{\text{Loc}}(z', z) \approx \text{Infl}_{\text{IP}}(z', z)$ as desired.

%% file: sections/appendix_b.tex
\section{ICP as Implicit Gradient Descent}\label{sec:appendix_icp_implicit}

This section outlines the construction in \citet{irie2022dualformneuralnetworks} and \cite{dai-etal-2023-gpt}, which connects the transformer attention head to an implicit update step on the in-context demonstration. Let $X_{z}, X_{z'} \in \mathbb{R}^{d_{in}}$ be the input representations of a training sample z and test sample z'. Furthermore, let $[X_{z}, X_{z'}]$ denote the concatenation of $X_{z}$ and $X_{z'}$. Then, the transformer attention mechanism can be expressed as

\begin{align}\label{eqn:attn}
    \text{Attention}(K, V, q) 
    &= W_{v} [X_{z}, X_{z'}] \text{Softmax}\left( \frac{(W_{k} [X_{z}, X_{z'}])^T q}{\sqrt{d_{in}}} \right) \notag \\
    &\approx W_{v} [X_{z}, X_{z'}] (W_{k} [X_{z}, X_{z'}])^T q \quad\quad \text{(i.e., linear attention)} \notag \\
    &= [W_{v}X_{z}, W_{v}X_{z'}] [W_{k}X_{z}, W_{k}X_{z'}]^T q \notag \\
    &= (W_{v}X_{z}(W_{k}X_{z})^T + W_{v}X_{z'}(W_{k}X_{z'})^T) q \notag \\
    &= W_{v} X_{z'} (W_{k} X_{z'})^T q + W_{v} X_{z} (W_{k} X_{z})^T q
\end{align}

\noindent Note that the attention head weights for $X_{z}$ (i.e., $W_{v} X_{z} (W_{k} X_{z})^T q$) in line \ref{eqn:attn} can be re-written as:
\begin{align}
    \text{LinearAttn}(W_{v} X_{z}, (W_{k} X_{z})^T, q) = W_{v} X_{z} (W_{k} X_{z})^T q
\end{align}

\noindent Furthermore, let $W_{z'} = W_{v} X_{z'} (W_{k} X_{z'})^T q$ denote the zero-shot ``attention head" weights that are obtained when we pass $X_{z'}$ through the model. We can now write Attention(K, V, q) as:
\begin{align}
    \text{Attention}(K, V, q) \approx W_{z'}q + \text{LinearAttn}(W_{v} X_{z}, (W_{k} X_{z})^T, q)
\end{align}

\noindent Next, \citet{irie-etal-2023-practical} showed the duality between  linear attention and linear layers optimized by gradient descent, such that:
\begin{align}
    \text{LinearAttn}(W_{v} X_{z}, (W_{k} X_{z})^T, q) \approx \Delta  W_{z} q
\end{align}

\noindent where $\Delta  W_{z}$ is the weight update after taking gradient descent step on $X_{z}$. Hence, we have: 
\begin{align}
    \text{Attention}(K, V, q) \approx (W_{z'} + \Delta  W_{z}) q
\end{align}

%% file: sections/appendix_c.tex
\newpage
\section{Additional Tables and Figures}\label{sec:appendix-tables-figures}
\input{tables/task_examples}
\input{tables/alpaca_eval}
\input{tables/instruction_examples}
\input{figures/task_similarity_appendix}
\input{figures/lin-func-input-task-corr-all}

%% file: tables/task_examples.tex
\begin{table*}[h!]
    \small
    \begin{center}
    \begin{tabular}{p{0.1\linewidth}|p{0.8\linewidth}}
    \toprule
    \textbf{Dataset} & \textbf{Example} \\
    \midrule
    Alpaca
    & \textbf{Instruction:} Create an array containing the first 10 prime numbers. \textbf{Response:} Here is an array containing the first 10 prime numbers: [2, 3, 5, 7, 11, 13, 17, 19, 23, 29]
    \\
    \midrule
    QA
    & \textbf{Question:} How old is the female main protagonist of Catching Fire? \textbf{Answer}: 16-years-old
    \\
    \midrule
    DocQA
    & \textbf{Question:} What monarch like accessory did he wear? \textbf{Context:}: When he became First Consul and later Emperor, Napoleon eschewed his general's uniform and habitually wore the simple green colonel uniform (non-Hussar)  ... \textbf{Answer}:  Iron Crown decorations
    \\
    \midrule
    Pretrain
    & A blog about my adventures as a professional garden writer, radio show host and obsessive veggie gardener! My first book, the award-winning, The Year Round Vegetable Gardener, was published in 2012 and Groundbreaking Food Gardens hit shelves in March 2014! 
    \\
    \bottomrule
    \end{tabular}
    \end{center}
    \caption{Demonstrations from the Alpaca, DocQA, QA, and pretrain datasets.}
    \label{tab:task-examples}
\end{table*}

%% file: tables/alpaca_eval.tex
\begin{table}[h!]
\centering
\begin{tabular}{c|c}
    \toprule
    \textbf{Subset} & \textbf{Number of Examples} \\
    \midrule
    Helpful Base \cite{bai2022training} & 129  \\
    Koala \cite{geng2023koala} &  156 \\
    Self-Instruct \cite{wang-etal-2023-self-instruct} & 252 \\
    Oasst &  188 \\
    Vicuna \cite{chiang2023vicuna} & 80 \\
    \bottomrule
    \end{tabular}
\caption{Breakdown of the subsets in the Alpaca Eval  \cite{li2023alpacaeval, dubois2024alpacafarm} dataset.}
\label{tab:alpaca-eval}
\end{table}

%% file: tables/instruction_examples.tex
\begin{table*}[ht!]
    \small
    \begin{center}
    \begin{tabular}{p{0.45\linewidth}|p{0.45\linewidth}}
    \toprule
    ICP & $\text{Infl}_{\text{IP}}$ \\
    \midrule
    \textbf{Instruction:} Create an array containing the first 10 prime numbers. \textbf{Response:} Here is an array containing the first 10 prime numbers: [2, 3, 5, 7, 11, 13, 17, 19, 23, 29]
    & 
    \textbf{Instruction:} Create an array containing the first 10 prime numbers. \textbf{Response:} Here is an array containing the first 10 prime numbers: [2, 3, 5, 7, 11, 13, 17, 19, 23, 29] \\
    \midrule
    \textbf{Instruction:} Write an algorithm for printing all prime numbers between 0 and 100. \textbf{Response:} Here is an algorithm for printing all prime numbers between 0 and 100: 1. Start with a variable `n` at 2, since 2 is the first prime number. 2. Create a loop that will iterate until `n` is greater than 100. 3. At each iteration of the loop, check if `n` is prime ...
    & 
    \textbf{Instruction:} Given an array of integers and a target number, find a triplet which adds up to the target number. Input: Array: [2, 5, 3, 6, 9], Target: 12 \textbf{Response:} One possible solution is to sort the array and then use two pointers to check for possible triplets. Here is the step-by-step process: 1. Sort the array in ascending order. In this case, it will be [2, 3, 5, 6, 9]. 2. Initialize a pointer `i` to the first element  ...
    \\
    \midrule
    \textbf{Instruction:} Generate a random string of 5 lowercase letters.
    \textbf{Response:} Here is a randomly generated string of 5 lowercase letters: "kdjyr"
    &
    \textbf{Instruction:} Create an algorithm to sort a given array of numbers. Input: [7, 1, 3, 5, 9] 
    \textbf{Response:} There are many algorithms to sort a given array of numbers. Below is one such algorithm using the Bubble Sort method: 1. Begin by comparing the first two elements of the array ...
    citep\\
    \midrule
    \textbf{Instruction:} Generate a random password with 8 characters, consisting of lower and upper-case letters and numbers. 
    \textbf{Response:} Here is a randomly generated password with 8 characters, consisting of lower and upper-case letters and numbers: "aB3dE6fG"
    &
    \textbf{Instruction:} Create an algorithm in pseudocode that sorts an array of numbers using the selection sort method. Input: Array: [3, 1, 4, 7, 8, 2]
    \textbf{Response:} Here is a pseudocode that sorts an array of numbers using the selection sort method: Procedure selection\_sort(Array: list of sortable items) BEGIN Declare index\_min, temporary  ...
    \\
    \midrule
    \textbf{Instruction:} Create an algorithm to find the maximum value in an array of numbers. Input: [3, 5, 7, 2]
    \textbf{Response:} Here is an algorithm to find the maximum value in an array of numbers: 1. Initialize the first element of the array as the maximum value. 2. Loop through the rest of the array, comparing each element with the current maximum value. 3. If the current element is ...
    citep
    &
    \textbf{Instruction:} Create a PHP script to detect whether a string contains a number.
    \textbf{Response:} Here is one way to create a PHP script to detect whether a string contains a number: ``` <?php function containsNumber(\$str) \{ // check if the input string contains any digits if (preg\_match('/\d/', \$str)) { return true; } else { return false; } \} // test ...
    \\
    \bottomrule
    \end{tabular}
    \end{center}
    \caption{The top-5 demonstrations selected by ICP and $\text{Infl}_{\text{IP}}$ for the test task: Instruction: Create an array containing the first 10 prime numbers.}
    \label{tab:instruction-examples}
\end{table*}

%% file: figures/task_similarity_appendix.tex
\begin{figure*}[h!]
\centering
\nocaption
   \begin{subfigure}{0.24\linewidth}
    \includegraphics[width=\linewidth]{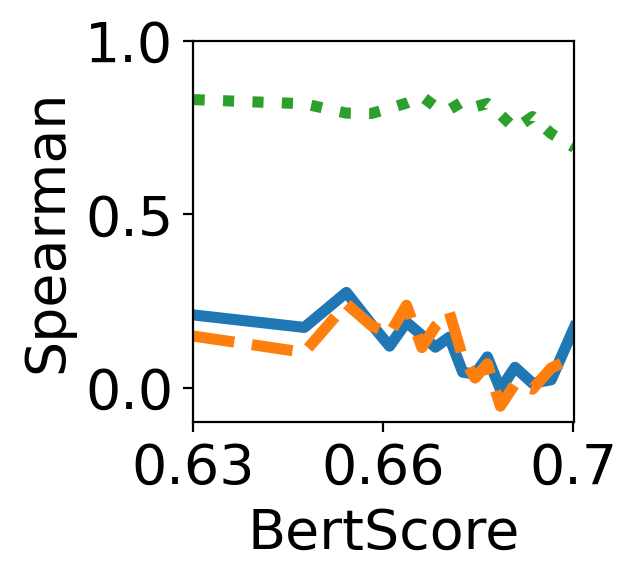}
    \caption{QA (Pythia)}
    \label{fig:bertscore_kmeans_qa_pythia}
  \end{subfigure}
  \begin{subfigure}{0.24\linewidth}
    \includegraphics[width=\linewidth]{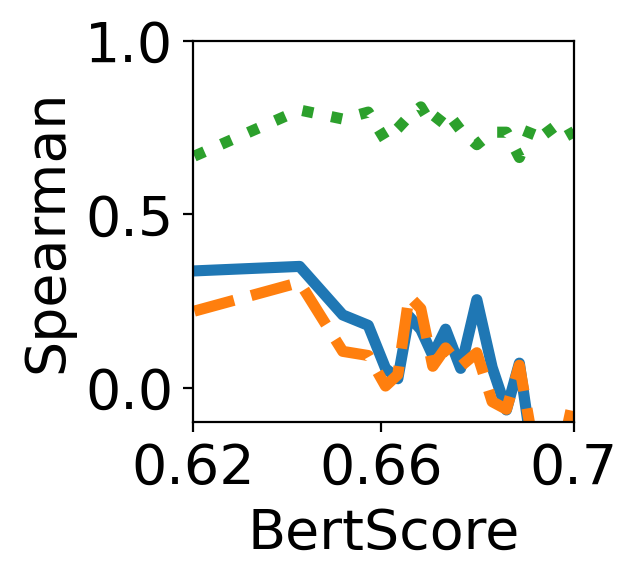}
    \caption{QA (Llama)}
    \label{fig:bertscore_kmeans_qa_llama}
  \end{subfigure}
  \caption{Correlation analysis between ICP, $\text{Infl}_{\text{Loc}}$, and $\text{Infl}_{\text{IP}}$ with respect to content similarity (BertScore) using test samples from Alpaca and training samples from QA dataset.}
  \label{fig:task-similarity-additional}
\end{figure*}

%% file: figures/lin-func-input-task-corr-all.tex
\begin{figure*}[th!]
  \centering
  \nocaption
  \begin{subfigure}{0.37\linewidth}
    \includegraphics[width=\linewidth]{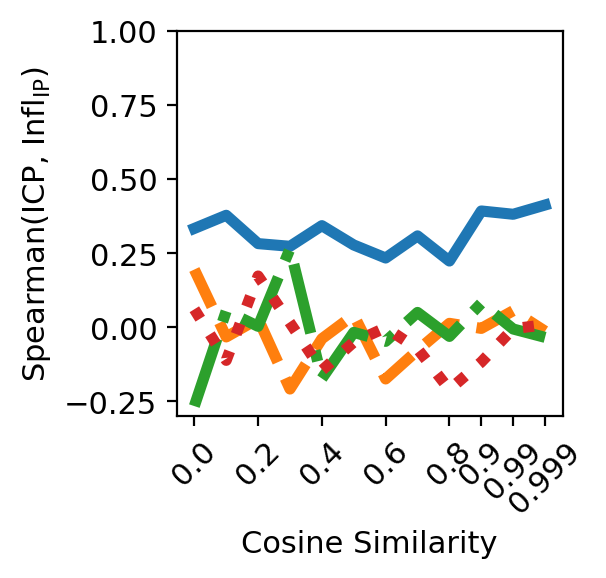}
    \caption{Task similarity}
    \label{fig:lin_func_task_icp_infl} 
  \end{subfigure}
  \begin{subfigure}{0.35\linewidth}
    \includegraphics[width=\linewidth]{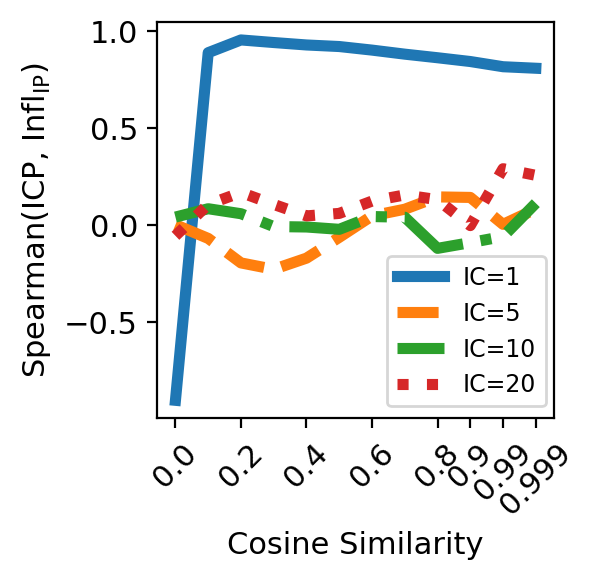}
    \caption{Content similarity}
    \label{fig:lin_func_input_icp_infl} 
  \end{subfigure}
  \caption{Correlation analysis between ICP and $\text{Infl}_{\text{IP}}$ as the task and content similarity of the ICL training demonstrations vary with respect to the test query. IC is the number of in-context training demonstrations used.}
  \label{fig:lin-func-input-task-corr-all}
\end{figure*}